\documentclass{article}
\usepackage{spconf,amsmath,graphicx}
\usepackage[pagebackref=true,breaklinks=true,bookmarks=false]{hyperref}
\usepackage[accsupp]{axessibility}
\usepackage{amsfonts}
\usepackage{float}
\usepackage{amssymb}
\usepackage{enumitem}
\usepackage{algorithm} 
\usepackage{algorithmic}
\usepackage[algo2e]{algorithm2e}
\usepackage{amsmath,amsthm,mathtools}
\usepackage{url}
\usepackage{stmaryrd}
\usepackage{cite}
\usepackage{multirow}
\usepackage[table,xcdraw]{xcolor}
\usepackage{url}
\usepackage{ragged2e}
\usepackage{epstopdf}
\usepackage{booktabs}
\usepackage{threeparttable}

\newcommand{\etal}{\textit{et al}. }
\newcommand{\ie}{\textit{i}.\textit{e}., }

\setlist{nolistsep}

\DeclareMathOperator{\diag}{\operatorname{diag}}

\DeclareMathOperator{\softmax}{\operatorname{softmax}}
\DeclareMathOperator{\relu}{\operatorname{ReLU}}

\newtheorem{theorem}{Theorem}

\newtheorem{definition}{Definition}

\title{Higher-order Sparse Convolutions in Graph Neural Networks}
\name{{Jhony H. Giraldo}$^{\star+}$, {Sajid Javed}$^{\ddagger}$, {Arif Mahmood}$^{\oast}$, {Fragkiskos D. Malliaros}$^{\dagger}$, {Thierry Bouwmans}$^{\S}$\thanks{${^+}$Corresponding author: \href{mailto:jhony.giraldo@telecom-paris.fr}{jhony.giraldo@telecom-paris.fr}}}

\address{$^\star$LTCI, Télécom Paris - Institut Polytechnique de Paris, France;\\
$^{\ddagger}$Khalifa University, United Arab Emirates; $^{\oast}$Information Technology University, Pakistan;\\
$^\dagger$Université Paris-Saclay, CentraleSupélec, Inria, Centre for Visual Computing (CVN), France; \\
$^{\S}$Laboratoire MIA, La Rochelle Université, France}

\begin{document}
%
\maketitle
\begin{abstract}
Graph Neural Networks (GNNs) have been applied to many problems in computer sciences.
Capturing higher-order relationships between nodes is crucial to increase the expressive power of GNNs.
However, existing methods to capture these relationships could be infeasible for large-scale graphs.
In this work, we introduce a new higher-order sparse convolution based on the Sobolev norm of graph signals.
Our Sparse Sobolev GNN (S-SobGNN) computes a cascade of filters on each layer with increasing Hadamard powers to get a more diverse set of functions, and then a linear combination layer weights the embeddings of each filter.
We evaluate S-SobGNN in several applications of semi-supervised learning.
S-SobGNN shows competitive performance in all applications as compared to several state-of-the-art methods.
\end{abstract}
\section{Introduction}
\label{sec:intro}

Graph representation learning and its applications have gained significant attention in recent years.
Notably, Graph Neural Networks (GNNs) have been extensively studied \cite{defferrard2016convolutional,kipf2017semi,velickovic2018graph,ioannidis2019recurrent,zhao2021distributed,pfrommer2021discriminability}.
GNNs extend the concepts of Convolutional Neural Networks (CNNs) \cite{lecun2015deep} to non-Euclidean data modeled as graphs.
GNNs have numerous applications like semi-supervised learning \cite{kipf2017semi}, graph clustering \cite{duval2022hoscpool}, point cloud semantic segmentation \cite{li2019deepgcns}, misinformation detection \cite{benamira2019fakenews}, and protein modeling  \cite{gainza2020deciphering}.
Similarly, other graph learning techniques have been recently applied to image and video processing applications \cite{egilmez2020graph, giraldo2020graph}.

Most GNNs update their node embeddings by computing specific operations in the neighborhood of each node.
This updating is limited when we want to capture higher-order vertex relationships between nodes.
Previous methods in GNNs have tried to capture these higher-order connections by taking powers of the sparse adjacency matrix \cite{frasca2020sign}, quickly converting this sparse representation into a dense matrix.
The densification of the adjacency matrix results in memory and scalability problems in GNNs.
Therefore, the use of these higher-order methods is limited for large-scale graphs.

In this work, we propose a new sparse GNN model that computes a cascade of higher-order filtering operations.
Our model is inspired by the Sobolev norm in Graph Signal Processing (GSP) \cite{pesenson2009variational,giraldo2022reconstruction}.
We modify the Sobolev norm using concepts of the Hadamard product between matrices to maintain the sparsity of the adjacency matrix.
We rely on spectral graph theory \cite{chung1997spectral} and the Schur product theorem \cite{horn2012matrix} to explain some mathematical properties of our filtering operation.
Our Sparse Sobolev GNN (S-SobGNN) employs a linear combination layer at the end of each cascade of filters to select the best power functions.
Thus, we improve expressiveness by computing a more diverse set of sparse graph-convolutional operations.
We evaluate S-SobGNN in semi-supervised learning tasks in several domains like tissue phenotyping in colon cancer histology images \cite{kather2016multi}, text classification of news \cite{lang1995newsweeder}, activity recognition with sensors \cite{anguita2013public}, and recognition of spoken letters \cite{fanty1991spoken}.

The main contributions of the current work are summarized as follows: 1) we propose a new GNN architecture that computes a cascade of higher-order filters inspired by the Sobolev norm in GSP, 2) some mathematical insights of S-SobGNN are introduced based on spectral graph theory \cite{chung1997spectral} and the Schur product theorem \cite{horn2012matrix}, and 3) we perform experimental evaluations on four publicly available benchmark datasets and compared S-SobGNN to seven GNN architectures.
Our algorithm shows the best performance against previous methods.
The rest of the paper is organized as follows.
Section \ref{sec:S-SobGNN} introduces the proposed GNN model.
Section \ref{sec:experiments_results} presents the experimental framework and results.
Finally, Section \ref{sec:conclusions} shows the conclusions.

\section{Sparse Sobolev Graph Neural Networks}
\label{sec:S-SobGNN}

\subsection{Preliminaries}

A graph is represented as $G=(\mathcal{V},\mathcal{E})$, where $\mathcal{V}=\{1,\dots,N\}$ is the set of $N$ nodes and $\mathcal{E}=\{(i,j)\}$ is the set of edges between nodes $i$ and $j$.
$\mathbf{A}\in \mathbb{R}^{N\times N}$ is the weighted adjacency matrix of the graph such that $\mathbf{A}(i,j)=a_{i,j}\in \mathbb{R}_+$ is the weight of the edge $(i,j)$, and $\mathbf{A}(i,j)=0~\forall~(i,j) \notin \mathcal{E}$.
As a result, $\mathbf{A}$ is symmetric for undirected graphs.
A graph signal is a function $x:\mathcal{V} \to \mathbb{R}$ and is represented as $\mathbf{x} \in \mathbb{R}^N$.
The degree matrix of $G$ is a diagonal matrix given by $\mathbf{D}=\diag(\mathbf{A1})$.
$\mathbf{L=D-A}$ is the combinatorial Laplacian matrix, and $\mathbf{\Delta=I-D}^{-\frac{1}{2}}\mathbf{AD}^{-\frac{1}{2}}$ is the symmetric normalized Laplacian \cite{ortega2018graph}.
The Laplacian matrix is a positive semi-definite matrix for undirected graphs with eigenvalues\footnote{$\lambda_N \leq 2$ in the case of the symmetric normalized Laplacian $\mathbf{\Delta}$.} $0=\lambda_1 \leq \lambda_2 \leq \dots \leq \lambda_N$ and corresponding eigenvectors $\{ \mathbf{u}_1,\mathbf{u}_2,\dots,\mathbf{u}_N\}$.
In GSP, the Graph Fourier Transform (GFT) of $\mathbf{x}$ is given by $\mathbf{\hat{x}}=\mathbf{U}^{{\mathsf{T}}}\mathbf{x}$, and the inverse GFT is $\mathbf{x} = \mathbf{U}\mathbf{\hat{x}}$ \cite{ortega2018graph}.
In this work, we use the spectral definitions of graphs to analyze our filtering operation.
However, the spectrum is not required for the implementation of S-SobGNN.

\subsection{Sobolev Norm}
\label{sec:sobolev_norm}

The Sobolev norm in GSP has been used as a regularization term to solve problems in 1) video processing \cite{giraldo2020graph,giraldo2020graphbgs}, 2) modeling of infectious diseases \cite{giraldo2020minimization}, and 3) interpolation of graph signals \cite{pesenson2009variational,giraldo2022reconstruction}.
\begin{definition}
    \label{dfn:sobolev_norm}
    For fixed parameters $\epsilon \geq 0$, $\rho\in \mathbb{R}$, the Sobolev norm is given by $\Vert \mathbf{x} \Vert_{\rho,\epsilon} \triangleq \Vert (\mathbf{L}+\epsilon \mathbf{I})^{\rho/2} \mathbf{x} \Vert$ \cite{pesenson2009variational}.
\end{definition}
\noindent When $\mathbf{L}$ is symmetric, we have that $\Vert \mathbf{x} \Vert_{\rho,\epsilon}^2$ is given by:
\begin{equation}
    \Vert \mathbf{x} \Vert_{\rho,\epsilon}^2 = \mathbf{x}^{\mathsf{T}}(\mathbf{L}+\epsilon\mathbf{I})^{\rho}\mathbf{x}.
    \label{eqn:sobolev_norm_rewritten}
\end{equation}

\noindent We divide the analysis of (\ref{eqn:sobolev_norm_rewritten}) into two parts: 1) when $\epsilon=0$, and 2) when $\rho=1$.
For $\epsilon=0$ in (\ref{eqn:sobolev_norm_rewritten}) we have:
\begin{equation}
    \mathbf{x}^{\mathsf{T}}\mathbf{L}^{\rho}\mathbf{x} = \mathbf{x}^{\mathsf{T}}\mathbf{U\Lambda}^{\rho}\mathbf{U}^{\mathsf{T}}\mathbf{x} = \hat{\mathbf{x}}^{\mathsf{T}}\mathbf{\Lambda}^{\rho}\hat{\mathbf{x}}=\sum_{i=1}^N \hat{\mathbf{x}}^2(i) \lambda_i^{\rho}.
    \label{eqn:penalized_laplacian}
\end{equation}
Notice that the spectral components $\hat{\mathbf{x}}(i)$ are penalized with powers of the eigenvalues $\lambda_i^{\rho}$ of $\mathbf{L}$.
Since the eigenvalues are ordered in increasing order, the higher frequencies of $\hat{\mathbf{x}}$ are penalized more than the lower frequencies when $\rho=1$, leading to a smooth function in $G$.
For $\rho>1$, the GFT $\hat{\mathbf{x}}$ is penalized with a more diverse set of eigenvalues.
We can have a similar analysis for the adjacency matrix $\mathbf{A}$ using the eigenvalue decomposition $\mathbf{A}^{\rho} = (\mathbf{V\Sigma V}^{\mathsf{H}})^{\rho}=\mathbf{V\Sigma}^\rho\mathbf{V}^{\mathsf{H}}$, where $\mathbf{V}$ is the matrix of eigenvectors, and $\mathbf{\Sigma}$ is the matrix of eigenvalues of $\mathbf{A}$.
In the case of $\mathbf{A}$, the GFT $\mathbf{\hat{x}}=\mathbf{V}^{{\mathsf{H}}}\mathbf{x}$.

\begin{figure*}
    \centering
    \includegraphics[width=0.82\textwidth]{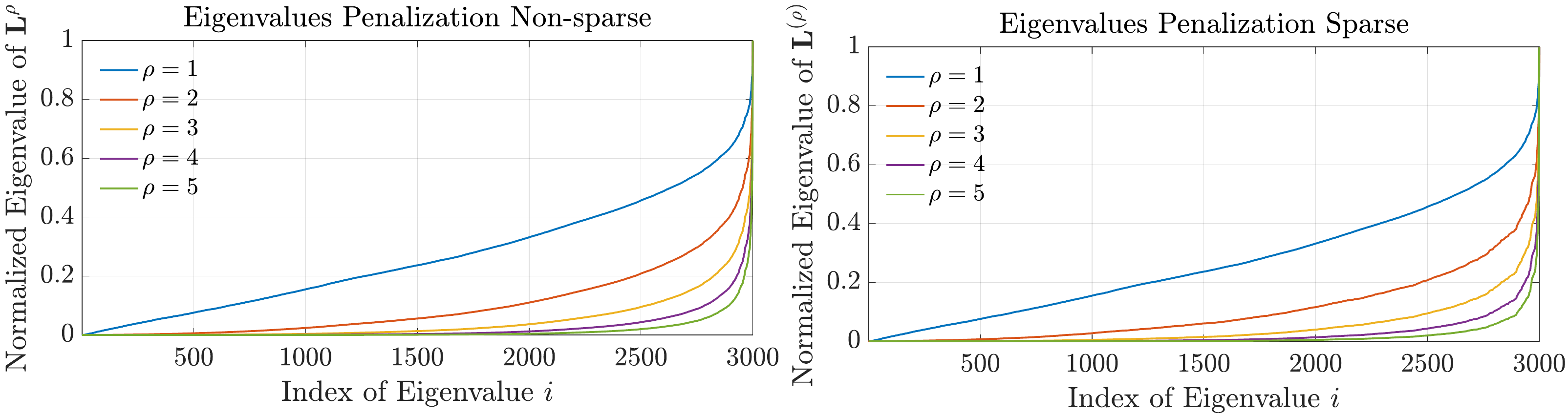}
    \caption{Eigenvalues penalization for the non-sparse and sparse matrix multiplications of the combinatorial Laplacian matrix.}
    \label{fig:spectrum_sparse_non_sparse}
\end{figure*}

For $\rho=1$ in (\ref{eqn:sobolev_norm_rewritten}) we have $\Vert \mathbf{x} \Vert_{\rho,\epsilon}^2 = \mathbf{x}^{\mathsf{T}}(\mathbf{L}+\epsilon\mathbf{I})\mathbf{x}$.
The term $(\mathbf{L}+\epsilon\mathbf{I})$ is associated with a better condition number\footnote{The condition number $\kappa(\mathbf{L})$ associated with the square matrix $\mathbf{L}$ is a measure of how well or ill-conditioned is the inversion of $\mathbf{L}$.} than using $\mathbf{L}$ alone.
For example, better condition numbers are associated with faster convergence rates in gradient descent methods as shown in \cite{giraldo2022reconstruction}.
For the Laplacian matrix $\mathbf{L}$, we know that $\kappa(\mathbf{L}) = \frac{\vert \lambda_{\text{max}}(\mathbf{L})\vert}{\vert \lambda_{\text{min}}(\mathbf{L})\vert} \approx \frac{\lambda_{\text{max}}(\mathbf{L})}{0} \rightarrow \infty$, where $\kappa(\mathbf{L})$ is the condition number of $\mathbf{L}$, $\lambda_{\text{max}}(\mathbf{L})$ is the maximum eigenvalue, and $\lambda_{\text{min}}(\mathbf{L})$ is the minimum eigenvalue of $\mathbf{L}$.
Since $\kappa(\mathbf{L}) \rightarrow \infty$, we have an ill-conditioned problem when relying on the Laplacian matrix alone.
On the other hand, for the Sobolev term, we have that $\mathbf{L}+\epsilon\mathbf{I} = \mathbf{U\Lambda U}^{\mathsf{T}}+\epsilon\mathbf{I} = \mathbf{U}(\mathbf{\Lambda}+\epsilon\mathbf{I})\mathbf{U}^{\mathsf{T}}$.
Therefore, $\lambda_{\text{min}}(\mathbf{L}+\epsilon\mathbf{I}) = \epsilon$, \ie $\mathbf{L}+\epsilon\mathbf{I}$ is positive definite ($\mathbf{L} +\epsilon\mathbf{I} \succ 0$) for $\epsilon > 0$, and:
\begin{equation}
    \kappa(\mathbf{L}+\epsilon\mathbf{I}) = \frac{\vert \lambda_{\text{max}}(\mathbf{L}+\epsilon\mathbf{I})\vert}{\vert \lambda_{\text{min}}(\mathbf{L}+\epsilon\mathbf{I})\vert} = \frac{\lambda_{\text{max}}(\mathbf{L})+\epsilon}{\epsilon} < \kappa(\mathbf{L})~\forall~\epsilon>0.
    \label{eqn:cond_number_sob_and_laplacian}
\end{equation}
Namely, $\mathbf{L}+\epsilon\mathbf{I}$ has a better condition number than $\mathbf{L}$.
It might not be evident why a better condition number could help in GNNs, where the inverses of the Laplacian or adjacency matrices are not required to perform the propagation rules.
However, some studies have indicated the adverse effects of bad-behaved matrices.
For example, Kipf and Welling \cite{kipf2017semi} used a renormalization trick ($\mathbf{A+I}$) in their filtering operation to avoid exploding/vanishing gradients.
Similarly, Wu \etal \cite{wu2019simplifying} showed that adding the identity matrix to $\mathbf{A}$ shrinks the graph spectral domain, resulting in a low-pass-type filter.

The previous theoretical analysis shows the benefits of the Sobolev norm about 1) the diverse frequencies computation in (\ref{eqn:penalized_laplacian}), and 2) the better condition number in (\ref{eqn:cond_number_sob_and_laplacian}).

\subsection{Sparse Sobolev Norm}
\label{sec:sparse_sob_norm}

The use of $\mathbf{L}$ or $\mathbf{A}$ in GNNs is computationally efficient because these matrices are usually sparse.
Therefore, we can perform a small number of sparse matrix operations.
For the Sobolev norm, the term $(\mathbf{L}+\epsilon\mathbf{I})^{\rho}$ can quickly become a dense matrix for large values of $\rho$, leading to scalability and memory problems.
To mitigate this limitation, we use a sparse Sobolev norm to keep the same sparsity level.
\begin{definition}
    \label{dfn:sparse_sobolev_term}
    Let $\mathbf{L}\in \mathbb{R}^{N\times N}$ be the Laplacian matrix of $G$.
    For fixed parameters $\epsilon \geq 0$ and $\rho \in \mathbb{N}$, the sparse Sobolev term for GNNs is introduced as the $\rho$ Hadamard multiplications of $(\mathbf{L}+\epsilon\mathbf{I})$ (the Hadamard power) such that:
    \begin{equation}
        (\mathbf{L}+\epsilon\mathbf{I})^{(\rho)}=(\mathbf{L}+\epsilon\mathbf{I}) \circ (\mathbf{L}+\epsilon\mathbf{I}) \circ \dots \circ (\mathbf{L}+\epsilon\mathbf{I}).
        \label{eqn:sparse_sobolev_term}
    \end{equation}
    For example, $(\mathbf{L}+\epsilon\mathbf{I})^{(2)}=(\mathbf{L}+\epsilon\mathbf{I}) \circ (\mathbf{L}+\epsilon\mathbf{I})$.
    Thus, the sparse Sobolev norm is given by: 
    \begin{equation}
        \Vert \mathbf{x} \Vert_{(\rho),\epsilon} \triangleq \Vert (\mathbf{L}+\epsilon \mathbf{I})^{(\rho/2)} \mathbf{x} \Vert.
        \label{eqn:sparse_sobolev_norm}
    \end{equation}
\end{definition}
Let $\langle \mathbf{x}, \mathbf{y} \rangle_{(\rho),\epsilon} = \mathbf{x}^{\mathsf{T}}(\mathbf{L}+\epsilon\mathbf{I})^{(\rho)}\mathbf{y}$ be the inner product between two graph signals $\mathbf{x}$ and $\mathbf{y}$ that induces the associated sparse Sobolev norm.
We can easily prove that the sparse Sobolev norm $\Vert \mathbf{x} \Vert_{(\rho),\epsilon} \triangleq \Vert (\mathbf{L}+\epsilon \mathbf{I})^{(\rho/2)} \mathbf{x} \Vert$ satisfies the basic properties of vector norms\footnote{We omit the proof due to space limitation.} for $\epsilon>0$ (for $\epsilon=0$ we have a semi-norm).
For the positive definiteness property, we need the Schur product theorem \cite{horn2012matrix}.

The sparse Sobolev term in (\ref{eqn:sparse_sobolev_term}) has the property of keeping the same sparsity level for any $\rho$.
Notice that $(\mathbf{L}+\epsilon\mathbf{I})^{\rho}$ is equal to the sparse Sobolev term if 1) we restrict $\rho$ to be in $\mathbb{N}$, and 2) we replace the matrix multiplication by the Hadamard product.
The theoretical properties of the Sobolev norm in (\ref{eqn:penalized_laplacian}) and (\ref{eqn:cond_number_sob_and_laplacian}) do not extend trivially to its sparse counterpart.
However, we can develop some theoretical insights using concepts of Kronecker products and the Schur product theorem \cite{horn2012matrix}.

\begin{theorem}
    \label{lem:spectrum_hadamard_product}
    Let $\mathbf{L}$ be any Laplacian matrix of a graph with eigenvalue decomposition $\mathbf{L}=\mathbf{U\Lambda U}^{\mathsf{T}}$, we have that:
    \begin{equation}
        \mathbf{L} \circ \mathbf{L} = \mathbf{L}^{(2)} = \mathbf{P}_N^{\mathsf{T}} (\mathbf{U} \otimes \mathbf{U})(\mathbf{\Lambda} \otimes \mathbf{\Lambda})(\mathbf{U}^{\mathsf{T}} \otimes \mathbf{U}^{\mathsf{T}})\mathbf{P}_N,
        \label{eqn:spectrum_hadamard_product}
    \end{equation}
    where $\mathbf{P}_N \in \{0,1\}^{N^2 \times N}$ is a partial permutation matrix.
    \begin{proof}
        For the spectral decomposition, we have that:
        \begin{equation}
            \mathbf{L}\otimes \mathbf{L}=(\mathbf{U} \otimes \mathbf{U})(\mathbf{\Lambda} \otimes \mathbf{\Lambda})(\mathbf{U}^{\mathsf{T}} \otimes \mathbf{U}^{\mathsf{T}}),
            \label{eqn:spectrum_kronecker_product}
        \end{equation}
        where we used the property of Kronecker products $(\mathbf{A} \otimes \mathbf{B})(\mathbf{C} \otimes \mathbf{D}) = \mathbf{AC} \otimes \mathbf{BD}$ \cite{horn2012matrix}.
        Similarly, we know that $\mathbf{S} \circ \mathbf{T} = \mathbf{P}_n^{\mathsf{T}}(\mathbf{S} \otimes \mathbf{T})\mathbf{P}_m$, where $\mathbf{S},\mathbf{T} \in \mathbb{R}^{n\times m}$, and $\mathbf{P}_n \in \{0,1\}^{n^2 \times n}, \mathbf{P}_m \in \{0,1\}^{m^2 \times m}$ are partial permutation matrices.
        If $\mathbf{S},\mathbf{T} \in \mathbb{R}^{n\times n}$ are square matrices, we have that $\mathbf{S} \circ \mathbf{T} = \mathbf{P}_n^{\mathsf{T}}(\mathbf{S} \otimes \mathbf{T})\mathbf{P}_n$ (Theorem 1 in \cite{visick2000quantitative}).
        We can then get a general form of the spectrum of the Hadamard product for $\rho=2$ using (\ref{eqn:spectrum_kronecker_product}) and Theorem 1 in \cite{visick2000quantitative} as follows: $\mathbf{L} \circ \mathbf{L} = \mathbf{L}^{(2)} = \mathbf{P}_N^{\mathsf{T}} (\mathbf{U} \otimes \mathbf{U})(\mathbf{\Lambda} \otimes \mathbf{\Lambda})(\mathbf{U}^{\mathsf{T}} \otimes \mathbf{U}^{\mathsf{T}})\mathbf{P}_N$.
    \end{proof}
\end{theorem}

Eq. (\ref{eqn:spectrum_hadamard_product}) is a closed-form solution regarding the spectrum of the Hadamard power for $\rho=2$.
Thus, the spectrum of the Hadamard multiplication is a compressed form of the Kronecker product of its spectral components.
The sparse Sobolev term we use in our S-SobGNN is given by $(\mathbf{L}+\epsilon\mathbf{I})^{(\rho)}$ so that the spectral components of the graph are changing for each value of $\rho$ as shown in \eqref{eqn:spectrum_hadamard_product}.

For the condition number of the Hadamard powers, we can use the Schur product theorem \cite{horn2012matrix}.
We know that $(\mathbf{L}+\epsilon\mathbf{I})^{(\rho)} \succ 0~\forall~\epsilon > 0$ since $(\mathbf{L}+\epsilon\mathbf{I}) \succ 0~\forall~\epsilon > 0$, and therefore $\kappa((\mathbf{L}+\epsilon\mathbf{I})^{(\rho)}) < \infty$.
For the adjacency matrix, the eigenvalues of $\mathbf{A}$ lie into $[-d,d]$, where $d$ is the maximal degree of $G$ \cite{nica2018spectral}.
Therefore, we can bound the eigenvalues of $\mathbf{A}$ into $[-1,1]$ by normalizing $\mathbf{A}$ such that $\mathbf{A}_N=\mathbf{D}^{-\frac{1}{2}}\mathbf{AD}^{-\frac{1}{2}}$.
As a result, we know that $\mathbf{A}_N+\epsilon \mathbf{I} \succ 0~\forall~\epsilon > 1$, and $(\mathbf{A}_N+\epsilon \mathbf{I})^{(\rho)}\succ 0~\forall~\epsilon > 1$. 
We can say that the theoretical developments of the sparse Sobolev norm hold to some extent the same developments of Section \ref{sec:sobolev_norm}, \ie a more diverse set of frequencies and a better condition number.
Figure \ref{fig:spectrum_sparse_non_sparse} shows five normalized eigenvalue penalizations for $\mathbf{L}^{\rho}$ (non-sparse) and $\mathbf{L}^{(\rho)}$ (sparse).
We notice that the normalized spectrum of $\mathbf{L}^{\rho}$ and $\mathbf{L}^{(\rho)}$ are very similar.
Finally, we should work with weighted graphs when using the adjacency matrix since $\mathbf{A}^{(\rho)} = \mathbf{A}~\forall~\rho \in \mathbb{N}$ for unweighted graphs.

\begin{figure*}
    \centering
    \includegraphics[width=0.96\textwidth]{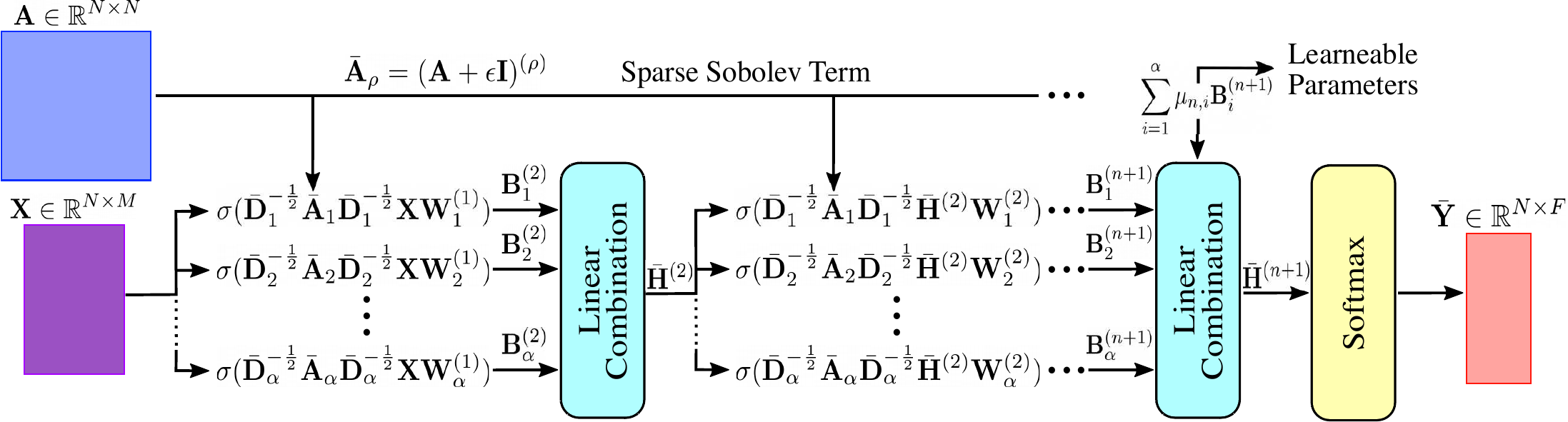}
    \caption{Basic configuration of our S-SobGNN architecture with $n$ layers and $\alpha$ filters per layer.}
    \label{fig:S-SobGNN_Architecture}
\end{figure*}

\subsection{Graph Neural Network Architecture}

Kipf and Welling \cite{kipf2017semi} proposed one of the most successful yet simple GNN, called Graph Convolutional Networks (GCNs):
\begin{equation}
    \mathbf{H}^{(l+1)}=\sigma(\tilde{\mathbf{D}}^{-\frac{1}{2}}\tilde{\mathbf{A}}\tilde{\mathbf{D}}^{-\frac{1}{2}}\mathbf{H}^{(l)}\mathbf{W}^{(l)}),
    \label{eqn:propagation_rule}
\end{equation}
where $\tilde{\mathbf{A}}=\mathbf{A}+\mathbf{I}$, $\tilde{\mathbf{D}}$ is the degree matrix of $\tilde{\mathbf{A}}$, $\mathbf{H}^{(l)}$ is the matrix of activations in layer $l$ such that $\mathbf{H}^{(1)}=\mathbf{X}$ (data matrix), $\mathbf{W}^{(l)}$ is the matrix of trainable weights in layer $l$, and $\sigma(\cdot)$ is an activation function.
The motivation of the propagation rule in (\ref{eqn:propagation_rule}) comes from the first-order approximation of localized spectral filters on graphs \cite{defferrard2016convolutional}.
Kipf and Welling \cite{kipf2017semi} used (\ref{eqn:propagation_rule}) to propose the vanilla GCN, which is composed of two graph convolutional layers as in \eqref{eqn:propagation_rule}. The first activation function is a Rectified Linear Unit ($\relu(\cdot)=\max(0,\cdot)$), and the final activation function is a softmax applied row-wise such that $\softmax(x_i)=\frac{1}{Q}\exp{(x_i)}$ where $Q=\sum_i \exp{(x_i)}$.
Finally, the vanilla GCN uses cross-entropy as a loss function.


We introduce a new filtering operation based on the sparse Sobolev term where our propagation rule is such that:
\begin{equation}
    \mathbf{B}_{\rho}^{(l+1)} = \sigma(\bar{\mathbf{D}}_{\rho}^{-\frac{1}{2}}\bar{\mathbf{A}}_{\rho}\bar{\mathbf{D}}_{\rho}^{-\frac{1}{2}}\bar{\mathbf{H}}^{(l)}\mathbf{W}_{\rho}^{(l)}),
    \label{eqn:propagation_rule_sobolev}
\end{equation}
where $\bar{\mathbf{A}}_{\rho}=(\mathbf{A}+\epsilon\mathbf{I})^{(\rho)}$ is the $\rho$th sparse Sobolev term of $\mathbf{A}$, $\bar{\mathbf{D}}_{\rho}$ is the degree matrix of $\bar{\mathbf{A}}_{\rho}$, and $\bar{\mathbf{H}}^{(1)}=\mathbf{X}$.
Notice that $\bar{\mathbf{A}}_{\rho}=\tilde{\mathbf{A}}$ when $\epsilon = 1$, and $\rho=1$, \ie our propagation rule is a generalization of the GCN model.
S-SobGNN computes a cascade of propagation rules as in (\ref{eqn:propagation_rule_sobolev}) with several values of $\rho$ in the set $\{1,2,\dots,\alpha\}$, and therefore a linear combination layer weights the outputs of each filter.
Figure \ref{fig:S-SobGNN_Architecture} shows the basic configuration of S-SobGNN.
Notice that our graph convolution is efficiently computed since the term $\bar{\mathbf{D}}_{\rho}^{-\frac{1}{2}}\bar{\mathbf{A}}_{\rho}\bar{\mathbf{D}}_{\rho}^{-\frac{1}{2}}~\forall~\rho \in \{1,2,\dots,\alpha\}$ is the same in all layers (so we can compute it offline), and also, these terms are sparse for any value of $\rho$ (given that $\mathbf{A}$ is also sparse).
S-SobGNN uses $\relu$ as the activation function for each filter, $\softmax$ at the end of the network, and the cross-entropy loss function.
The basic configuration of S-SobGNN is defined by the number of filters $\alpha$ in each layer, the parameter $\epsilon$, the number of hidden units of each $\mathbf{W}_{\rho}^{(l)}$, and the number of layers $n$.
When we construct weighted graphs with Gaussian kernels, the weights of the edges are in the interval $[0,1]$.
As a consequence, large values of $\rho$ could make $\bar{\mathbf{A}}_{\rho}=\mathbf{0}$, and the diagonal elements of $\bar{\mathbf{D}}_{\rho}^{-\frac{1}{2}}$ could become $\infty$.
Similarly, large values of $\alpha$ make very wide architectures with a high parameter budget, so it is desirable to maintain a reasonable value for $\alpha$.
The computational complexity of S-SobGNN is $\mathcal{O}(n \alpha \vert \mathcal{E} \vert + n \alpha)$.
For comparison, the computational complexity of a $n$-layers GCN is $\mathcal{O}(n \vert \mathcal{E} \vert)$.
The exact complexity of both methods also depends on the feature dimension, the hidden units, and the number of nodes in the graph, which we omit for simplicity.

\section{Experiments and Results}
\label{sec:experiments_results}

S-SobGNN is compared to eight GNN architectures: Chebyshev filters (Cheby) \cite{defferrard2016convolutional}, GCN \cite{kipf2017semi}, GAT \cite{velickovic2018graph}, SIGN \cite{frasca2020sign}, SGC \cite{wu2019simplifying}, ClusterGCN \cite{chiang2019cluster}, SuperGAT \cite{kim2021find}, and Transformers \cite{shi2021masked}.
We test S-SobGNN in several semi-supervised learning tasks including, cancer detection in images \cite{lang1995newsweeder}, text classification of news (20News) \cite{lang1995newsweeder}, Human Activity Recognition using sensors (HAR)\cite{anguita2013public}, and recognition of isolated spoken letters (Isolet) \cite{fanty1991spoken}.
We frame the semi-supervised learning problem as a node classification task in graphs, where we construct the graphs with a $k$-Nearest Neighbors ($k$-NN) method and a Gaussian kernel with $k=30$.
We split the data into train/validation/test sets with $10\%$/$45\%$/$45\%$, where we first divide the data into a development set and a test set.
This is done once to avoid using the test set in the hyperparameter optimization.
We tune the hyperparameters of each GNN with a random search with $100$ repetitions and five different seeds for the validation set.
We report average accuracies on the test set using $50$ different seeds with $95\%$ confidence intervals calculated by bootstrapping with $1,000$ samples.
Table \ref{tbl:results_classification} shows the experimental results.
S-SobGNN shows the best performance against state-of-the-art methods.

\begin{table}[]
\centering
\caption{Accuracy (in \%) for the baseline methods and our S-SobGNN algorithm in four datasets for semi-supervised learning, inferring the graphs with a $k$-NN method.}
\label{tbl:results_classification}
\begin{threeparttable}
\makebox[\linewidth]{
\scalebox{0.708}{
\begin{tabular}{r|cccc}
\toprule
\textbf{Model} & \textbf{Cancer} & \textbf{20News} & \textbf{HAR} & \textbf{Isolet} \\
\midrule
Cheby \cite{defferrard2016convolutional} & $87.55 \pm 3.91$ & $70.36 \pm 1.14$ & $73.14 \pm 7.01$ & $69.70 \pm 1.47$ \\
GCN \cite{kipf2017semi} & $76.71 \pm 4.47$ & $51.76 \pm 2.11$ & $66.26 \pm 4.91$ & $55.55 \pm 2.72$ \\
GAT \cite{velickovic2018graph} & $73.51 \pm 4.87$ & $48.72 \pm 2.21$ & $59.13 \pm 6.30$ & $60.00 \pm 2.21$ \\
SIGN \cite{frasca2020sign} & \color{blue} $\textbf{\textit{\underline{89.55}}} \pm \textbf{\textit{\underline{0.38}}}$ & \color{blue} $\textbf{\textit{\underline{71.79}}} \pm \textbf{\textit{\underline{0.25}}}$ & \color{blue} $\textbf{\textit{\underline{90.98}}} \pm \textbf{\textit{\underline{0.25}}}$ & \color{blue} $\textbf{\textit{\underline{84.02}}} \pm \textbf{\textit{\underline{0.30}}}$ \\
SGC \cite{wu2019simplifying} & $72.80 \pm 4.71$ & $54.68 \pm 1.99$ & $42.19 \pm 3.44$ & $41.55 \pm 1.91$ \\
ClusterGCN \cite{chiang2019cluster} & $74.45 \pm 5.37$ & $60.56 \pm 2.18$ & $57.70 \pm 5.48$ & $63.99 \pm 2.24$ \\
SuperGAT \cite{kim2021find} & $70.56 \pm 5.14$ & $57.52 \pm 1.93$ & $56.04 \pm 5.32$ & $58.49 \pm 2.27$ \\
Transformer \cite{shi2021masked} & $71.10 \pm 5.45$ & $57.48 \pm 2.29$ & $66.01 \pm 6.10$ & $66.24 \pm 2.39$ \\
\midrule
S-SobGNN (ours) & \color{red} $\textbf{93.11} \pm \textbf{0.45}$ & \color{red} $\textbf{72.18} \pm \textbf{0.32}$ & \color{red} $\textbf{92.85} \pm \textbf{0.58}$ & \color{red} $\textbf{86.17} \pm \textbf{0.34}$ \\
\bottomrule
\end{tabular}
}
}
\begin{tablenotes}\footnotesize
\item \scriptsize \hspace{0.15cm} The best and second-best performing methods on each dataset are shown in {\color{red}\textbf{red}}\\and {\color{blue}\textbf{\underline{\textit{blue}}}}, respectively.
\end{tablenotes}
\end{threeparttable}
\end{table}

\section{Conclusions}
\label{sec:conclusions}

In this work, we extended the concept of Sobolev norms using the Hadamard product between matrices to keep the sparsity level of the graph representations.
We introduced a new Sparse GNN architecture using the proposed sparse Sobolev norm.
Similarly, certain theoretical notions of our filtering operation were provided in Sections \ref{sec:sobolev_norm} and \ref{sec:sparse_sob_norm}.
Finally, S-SobGNN outperformed several methods of the literature in four semi-supervised learning tasks.\\

\noindent \textbf{Acknowledgments:} This work was supported by the DATAIA Institute as part of the ``Programme d'Investissement d'Avenir'', (ANR-17-CONV-0003) operated by CentraleSupélec, and by ANR (French National Research Agency) under the JCJC project GraphIA (ANR-20-CE23-0009-01).

\bibliographystyle{IEEEbib}
\bibliography{bibfile}

\end{document}